\documentclass[sigconf]{acmart}

\usepackage{tikz}
\usetikzlibrary{positioning, arrows.meta, calc, fit, backgrounds}
\usepackage{booktabs}
\usepackage{enumitem}
\usepackage{amsmath}

\copyrightyear{2026}
\acmYear{2026}
\setcopyright{cc}
\setcctype{by}
\acmConference[RecSys '26]{20th ACM Conference on Recommender Systems}{September 27-October 02, 2026}{Minneapolis, MN, USA}
\acmBooktitle{20th ACM Conference on Recommender Systems (RecSys '26), September 27-October 02, 2026, Minneapolis, MN, USA}
\acmDOI{10.1145/3773078.3831875}
\acmISBN{979-8-4007-2284-4/2026/09}

\begin{document}
\raggedbottom

\title{DRanker: A Transformer-Based Multi-Task Ranking Model for the Disney+ Homepage}

\author{Priya Nirmal Singh Khokher}
\orcid{0009-0008-3228-2627}
\email{priya.nirmal.singh.khokher@disney.com}
\affiliation{%
  \institution{The Walt Disney Company}
  \country{United States}
}

\author{Daniel Nemirovsky}
\orcid{0000-0003-4488-4752}
\email{daniel.nemirovsky@disney.com}
\affiliation{%
  \institution{The Walt Disney Company}
  \country{United States}
}

\author{Benjamin P. Whitesell}
\orcid{0009-0004-5420-5551}
\email{benjamin.whitesell@disney.com}
\affiliation{%
  \institution{The Walt Disney Company}
  \country{United States}
}

\author{Ethan Sukrae Lee}
\orcid{0000-0002-0873-9526}
\email{ethan.lee2@disney.com}
\affiliation{%
  \institution{The Walt Disney Company}
  \country{United States}
}

\author{Eric Wulf}
\orcid{0009-0003-0471-3518}
\email{eric.wulf@disney.com}
\affiliation{%
  \institution{The Walt Disney Company}
  \country{United States}
}

\author{Tom Chee-Duffy}
\email{thomas.duffy@disney.com}
\affiliation{%
  \institution{The Walt Disney Company}
  \country{United States}
}

\author{Nicholas Flores}
\email{nicholas.flores@disney.com}
\affiliation{%
  \institution{The Walt Disney Company}
  \country{United States}
}

\renewcommand{\shortauthors}{Khokher et al.}

\begin{abstract}
We present DRanker (Disney+ Ranker), a transformer-based multi-task ranking model serving personalized recommendations on the Disney+ homepage. We describe the encoder-only architecture over enriched engagement-history sequences paired with a lightweight causal debiasing mechanism, and the inference stack that enables transformer-based ranking at scale. Through iterative development, we systematically evaluated more complex alternatives and found that simpler designs consistently outperformed them on offline metrics, qualitative behavioral evaluation, and in production. Deployed globally, DRanker drives over 1\% lift in our key engagement metric and 0.5\% lift in daily active users across all experiences powered by the ranker. We share practical lessons on data engineering, debiasing, and architectural choices that shaped our ranking system at scale.
\end{abstract}



\maketitle

\section{Introduction}

Prior to DRanker (Disney+ Ranker), the Disney+ homepage was powered by several fragmented ranking models, each occupying dedicated real estate on the homepage with their own region-specific support or fallback logic. While each model was sophisticated in its own right, they were siloed: designed to offset each other's failure modes without sharing a unified view of the user. A common challenge specific to Disney+ is that a single profile often contains engagement from both children and adults. Some models overindexed on popularity bias for kids content and rewatches, while others prioritized general entertainment series or recent trending titles. Yet they all drew from the same underlying signal, user engagement history, with each applying its own sampling strategy, training pipeline, and update cadence. The result was a collection of siloed experts. None possessed holistic knowledge of user or region preferences. Collectively they were expensive to maintain, each requiring dedicated training data pipelines and batch storage for precomputed recommendations, and as we iterated on these models independently their goals diverged rather than converging toward a unified product experience.

This fragmented paradigm stands in contrast to foundational model approaches \cite{bommasani2021foundation} that leverage advances in sequential modelling: transformer architectures that learn from longer engagement histories spanning regions, devices, and temporal contexts. However, complex research architectures rarely translate cleanly into production systems \cite{dacrema2019progress, sculley2014tech_debt}. To make a meaningful impact, we needed calculated, incremental moves that could outperform the multi-model status quo.

The goal was a unified global ranker: a single model free of bespoke training data curation logic, served online rather than in batch (eliminating the cost of storing precomputed recommendations), and providing a stable foundation for adding new recommendation objectives. This goal introduced new functional requirements. The system had to meet strict latency constraints, operate without requiring additional infrastructure resources, and, above all, deliver measurable production gains. From these challenges, the unified ranker was born. We call it DRanker. Through its development, we repeatedly found that simpler, well-motivated designs outperformed more complex alternatives.

\textbf{Contributions.} This paper makes the following contributions:

\begin{enumerate}[leftmargin=*,nosep]
\item We describe the design and deployment of DRanker, a transformer-based multi-task ranking model serving the Disney+ homepage globally.
\item We present a causal debiasing tower \cite{cam2_2021, dice2021} that controls for popularity and conformity bias using only the transformer's own representations, without requiring additional data sources, and show it is more effective and lower-effort than alternatives like IPS weighting or weighted loss functions.
\item We share a systematic comparison of architectural choices (gated expert networks, deep cross networks, cross-attention, projection-based pooling) and explain why simpler designs won in production.
\item We detail practical data-engineering decisions (first-interaction sampling, stratified training data, and random negatives) that had outsized impact on ranking quality at scale.
\item We discuss DRanker's real-time serving layer, including LLM-based model promotion tests that evaluate recommendation quality against synthetic user profiles every day.
\end{enumerate}

\section{System Overview}

The Disney+ content-discovery pipeline follows a standard retrieve-then-rank architecture. Upstream retrieval systems generate candidate sets of movies and series for each user profile. DRanker acts as the ranking layer: given a user's engagement history and current session context, it scores every candidate title and produces a personalized ordering that surfaces titles on the homepage.

DRanker's scope is the ranking stage \cite{youtube2016}. For each candidate title, it must produce an engagement score for the title and an expected length of engagement. These two objectives are the output heads that define DRanker's multi-task formulation \cite{mtl_survey2023}.

\section{Training Data Construction}

\begin{figure}[t]
\centering
\resizebox{\columnwidth}{!}{%
\begin{tikzpicture}[
    >={Stealth[length=3pt]},
    funnel/.style={draw, very thick, fill=#1, minimum height=1.0cm, text centered, font=\small},
]
\fill[fill=blue!8, draw=black!70, very thick, rounded corners=2pt]
    (-5.0, 0) -- (5.0, 0) -- (5.0, -1.0) -- (-5.0, -1.0) -- cycle;
\node[font=\small] at (0, -0.5) {All Engagement};
\fill[fill=blue!12, draw=black!70, very thick, rounded corners=2pt]
    (-4.2, -1.4) -- (4.2, -1.4) -- (4.2, -2.4) -- (-4.2, -2.4) -- cycle;
\node[font=\small] at (0, -1.9) {Qualified by Engagement Length};
\fill[fill=blue!18, draw=black!70, very thick, rounded corners=2pt]
    (-3.4, -2.8) -- (3.4, -2.8) -- (3.4, -3.8) -- (-3.4, -3.8) -- cycle;
\node[font=\small] at (0, -3.3) {First-Time Interactions};
\fill[fill=blue!25, draw=black!70, very thick, rounded corners=2pt]
    (-2.6, -4.2) -- (2.6, -4.2) -- (2.6, -5.2) -- (-2.6, -5.2) -- cycle;
\node[font=\small\bfseries] at (0, -4.7) {Stratified Sampling};
\node[font=\small, anchor=west] at (5.5, -0.5) {All platform engagement};
\node[font=\small, anchor=west] at (5.5, -1.9) {Filter accidental engagement};
\node[font=\small, anchor=west] at (5.5, -3.3) {Discovery signal only};
\node[font=\small, anchor=west] at (5.5, -4.7) {Training samples};
\draw[very thick, black!70] (-5.0, -1.0) -- (-4.2, -1.4);
\draw[very thick, black!70] (5.0, -1.0) -- (4.2, -1.4);
\draw[very thick, black!70] (-4.2, -2.4) -- (-3.4, -2.8);
\draw[very thick, black!70] (4.2, -2.4) -- (3.4, -2.8);
\draw[very thick, black!70] (-3.4, -3.8) -- (-2.6, -4.2);
\draw[very thick, black!70] (3.4, -3.8) -- (2.6, -4.2);
\end{tikzpicture}%
}
\caption{Training data construction funnel. Each stage progressively filters and refines the raw viewing signal into high-quality training samples.}
\Description{Four-stage funnel diagram. Top layer: All Engagement. Second: Qualified by Engagement Length. Third: First-Time Interactions. Bottom: Stratified Sampling. Each stage is narrower than the previous, with annotations on the right explaining the filtering rationale.}
\label{fig:data-funnel}
\end{figure}
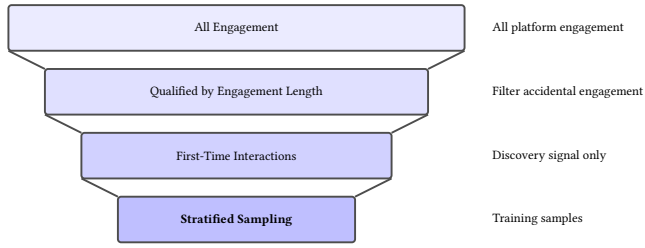

Models are retrained frequently during the week on a large-scale commodity GPU instance. Training data is generated via a daily ETL pipeline that samples all qualified interactions from the Disney+ platform, retaining engagement history over several months and producing hundreds of millions of training samples exported as Mosaic Data Shards.

Data engineering decisions had an outsized impact on DRanker's ranking quality. We describe the key choices and their rationale.

\subsection{Positive Sample Selection}

We define a positive interaction as a qualified engagement with the title, filtering out accidental or negligible interactions. From the full stream of viewing activity, we apply two filtering steps:

\textbf{First-interaction only.} For each (profile, title) pair, we retain only the first engagement. Continue-watching and re-watching are removed. The rationale is that ranking on the homepage is fundamentally a \textit{discovery} task \cite{othertube2022}: we are predicting what a user will engage with next given their current history. Continue-watching does not update a user's preference/history state towards that end. This filter also reduces the training set size substantially, sharpening the signal for the ranking objective.

\textbf{Stratified sampling.} Naive sampling of viewing activity is heavily skewed toward popular titles and high-activity regions. To ensure diversity, we stratify positive samples across different categories such as title types, regions, and devices. This ensures that minority titles and underrepresented regions are present in training data, which is critical for a globally deployed model.

\subsection{Negative Sampling}

For each positive training example, we generate a set of negative examples by randomly sampling titles from the content catalog. While more sophisticated negative-sampling strategies exist, including impression-based negatives (titles shown but not clicked) and dissimilar negatives, we found random sampling to be a strong baseline for several reasons:

\begin{itemize}
\item \textbf{Boundary placement:} Effective negatives should sit near the decision boundary between positive and negative classes. Our experiments showed that dissimilar negatives (selected to be far from the user's interests) performed \textit{worse} than random, because they are trivially distinguishable and provide weak gradient signal.
\item \textbf{Computational cost:} Impression-based negatives require joining interaction logs with impression logs at scale, which was prohibitively expensive for our initial delivery timeline.
\item \textbf{Empirical strength:} Random negatives provided a robust baseline that our ablations consistently validated.
\end{itemize}

\subsection{Sequence Construction}

Each training example includes the user's engagement history as a short window of recent engagement, ordered chronologically. Each element in the sequence is enriched with contextual information from when the title was engaged:

\begin{itemize}
\item Content embedding (LLM-derived text embedding and learned ID-based embedding of the title)
\item Contextual features such as device type and temporal signals
\item Engagement features such as engagement length.
\end{itemize}

A strict temporal constraint is enforced: the engagement-history sequence contains only titles engaged \textit{before} the candidate title, preventing label leakage.

\subsection{Candidate and Contextual Features}

The candidate title is represented as the concatenation of an LLM-derived text embedding and a learned ID-based embedding. Contextual features are encoded via learned ID-based embeddings shared with the analogous features in the engagement-history sequence. Temporal contextual features are cyclically encoded.

\section{Model Architecture}

\begin{figure*}[t]
\centering
\includegraphics[width=0.85\textwidth]{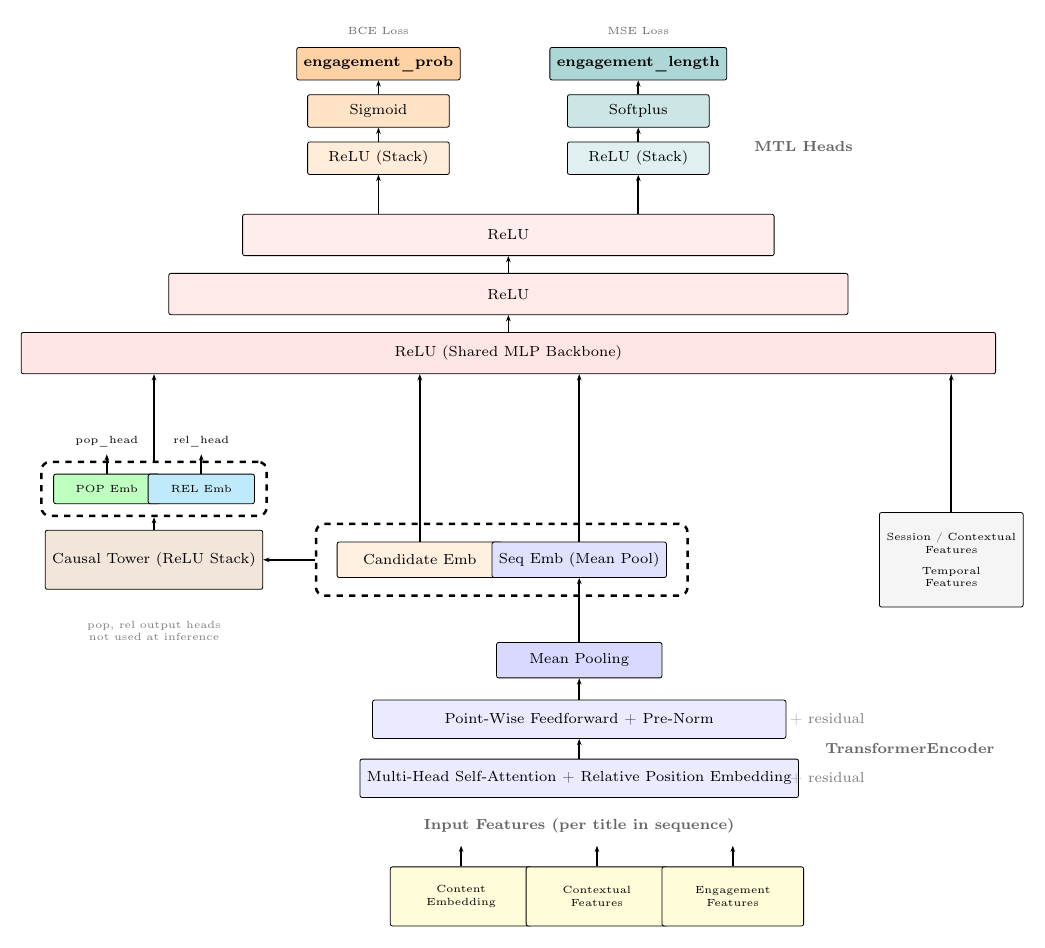}
\caption{DRanker model architecture. The transformer encoder processes enriched engagement-history sequences. The shared MLP backbone receives embeddings from the sequential encoder, candidate item, contextual features, and causal tower. Two task-specific heads predict engagement probability and expected engagement length. The causal tower's pop/rel output heads are discarded at inference; only its learned embeddings are retained.}
\Description{Model architecture diagram. Bottom: input features per sequence title (content embedding, contextual features, engagement features) feed into a transformer encoder with self-attention and mean pooling. Middle: candidate embedding, sequence embedding, session/contextual features, and causal tower embeddings feed into a shared ReLU MLP backbone. Top: two task-specific heads output engagement probability and engagement length. A separate causal tower on the left produces popularity and relevance embeddings with gradient isolation.}
\label{fig:architecture}
\end{figure*}

While the recommender-systems literature offers sophisticated multi-task learning (MTL) heads \cite{ple2020, mtl_survey2023} and gating mechanisms, we invested primarily in building a strong \textit{base} model. Our early experiments with averaged engagement-history embeddings showed consistently poor performance as history length grew beyond a few months. This motivated a shift to sequential modeling via a transformer encoder \cite{vaswani2017}.

\subsection{Transformer Encoder}

\textbf{Input representation.} Each item in the engagement-history sequence is represented as the concatenation of: (a) an LLM-derived text embedding of the title's metadata (synopsis, genre, cast), (b) a learned ID-based embedding, (c) a learned device embedding, (d) temporal features (cyclically encoded), and (e) the engagement length. This enriched representation allows the transformer to attend not just to \textit{what} the user engaged with, but \textit{when}, \textit{where} (device), and \textit{how much} they engaged.

\textbf{Self-attention with relative position embeddings (RPE).} The encoder uses multi-head self-attention \cite{sasrec2018} to produce contextualized representations of each item in the sequence. We used bidirectional attention \cite{bert2019, bert4rec2019} which is well suited to recommendations, to produce contextualized profile embeddings. We experimented with absolute position embeddings, learned position embeddings, and relative position embeddings \cite{t5_2020}. RPE consistently outperformed the alternatives. Because RPE are computed as biases added to attention scores within the multi-head self-attention mechanism (rather than as standalone embeddings added to the input), they naturally capture the \textit{relative} temporal ordering of items without being tied to absolute positions. This is important for a ranking model where sequences of varying lengths are processed. Beyond ordering, RPE captures relative shifts in user taste over time. For instance, recognizing that a user who recently engaged with \textit{Frozen} is likely to engage with \textit{Frozen 2}, regardless of where these titles fall in the absolute sequence. This property also benefits shared profiles where multiple household members contribute to the same engagement history (e.g., children watching animated content in the morning and adults watching thrillers at night) - a recognized failure mode of our then multi-ranker system. RPE allows the model to learn these recurring temporal patterns without conflating them with fixed sequence positions. Critically, our decision to adopt RPE was not driven by offline metrics alone. During qualitative inspection of mock-profile test sets, we observed that RPE-based models correctly predicted sequels for engagement history with prequels (a critical capability given our catalog's depth of franchise content) within substantially fewer training epochs than alternatives. Seeing the model produce recommendations that matched our intuitive expectations early in training gave us confidence that RPE was learning the right inductive biases, not just optimizing the right loss.

\textbf{Pointwise feedforward and mean pooling.} The output of the multi-head self-attention layer passes through a pointwise feedforward network with pre-norm residual connections. The resulting per-item representations are mean-pooled to produce a single profile-history embedding that maps diverse engagement histories into a common latent space.

\subsection{Multi-Task Learning Head}

DRanker produces two outputs via a multi-task learning head with \textit{hard parameter sharing}:

\begin{enumerate}
\item \textbf{engagement\_prob}: the probability the user will engage with the candidate title, optimized via binary cross-entropy (BCE) loss:

$$\mathcal{L}_{\text{BCE}} = -\frac{1}{N}\sum_{i=1}^{N} \left[ y_i \log(\hat{p}_i) + (1 - y_i) \log(1 - \hat{p}_i) \right]$$

where $y_i \in \{0, 1\}$ indicates whether the user engaged with the title and $\hat{p}_i$ is the predicted engagement probability. We modelled ranking as a classification problem following a \textit{pointwise ranking} paradigm \cite{liu2009learning, li2011short}. Random negative sampling (Section 3.2) means this output functions as a ranking score rather than a calibrated probability: it reflects whether a title fits the user's preferences, not a calibrated impression-to-engagement rate.

\item \textbf{engagement\_length}: the predicted engagement length, optimized via mean squared error (MSE) loss.
\end{enumerate}

The input to the MTL head is the concatenation of: the mean-pooled transformer output (sequential profile embedding), an average embedding of the user's full engagement history, the candidate item's LLM-derived text embedding, and contextual and temporal features. In configurations that include the causal debiasing tower (Section 4.3), its learned embeddings are appended to this input.

Both heads share a common stack of ReLU layers (the ``backbone''), from which task-specific ReLU stubs branch off. This follows the YouTube ranking architecture \cite{youtube2016}.

\textbf{Inference score.} At inference time, we experimented with variants of the ranking score: using the engagement\_prob head alone, and a combination of both heads. We discuss the results of these variants in Section 5.

\subsection{Causal Debiasing Tower}

\subsubsection{Motivation.} Despite strong offline precision and recall, qualitative evaluation of early DRanker models, with all previously described architectural choices in place, revealed a pronounced \textit{popularity bias} \cite{mansoury2020, chaney2018}: the ranker would surface a small set of globally popular or trending titles to nearly every profile. These recommendations stood out in qualitative testing not because popular titles are inherently unwanted (popularity can be a signal of quality) but because they were irrelevant to the user's engagement history. The issue is not recommending popular content; it is recommending popular content \textit{indiscriminately}, a known failure mode across the prior multi-model ecosystem that no individual ranker had fully resolved.

Popularity bias in recommendation is well-studied \cite{gao2022_causal_survey, abdollahpouri2019}, and while post-ranking methods exist \cite{abdollahpouri2019, bugliarello2022mostra}, we wanted to adopt strategies that are corrective in the ranker itself. We explored two mitigation strategies:

\begin{itemize}
\item \textbf{IPS weighting / weighted logistic regression:} Re-weight training samples by the inverse of their exposure probability to correct for selection bias \cite{schnabel2016}. Under this formulation, the BCE loss becomes:

$$\mathcal{L}_{\text{W-BCE}} = -\frac{1}{N}\sum_{i=1}^{N} w_i \left[ y_i \log(\hat{p}_i) + (1 - y_i) \log(1 - \hat{p}_i) \right]$$

where $w_i = \frac{|S|}{|K| \cdot |S_k|}$, with $|S|$ total samples, $|K|$ number of strata, and $|S_k|$ samples in stratum $k$, giving minority titles higher weight during training.

\item \textbf{Causal embeddings / disentangled representations \cite{gao2022_causal_survey}:} Learn separate representations for popularity/conformity effects and genuine user interest, then discard the bias components at inference.
\end{itemize}

We initially explored weighted logistic regression as a debiasing approach. However, we encountered a practical challenge: the loss was highly unstable on our stratified training data. Because our stratification ensures representation of minority titles and regions with fewer subscribers than others, individual mini-batches can have high variance in sample weights, leading to erratic loss curves. Several experiments were conducted around tuning the weighting scheme to stabilize training (smoothing the weights, explicit per-sample weight calibration) but added substantial engineering effort with no clear path to convergence. Additionally, IPS weights are known to drift in production \cite{zhao2019_youtube_multitask} providing lesser confidence.

Instead, we turned to the causal-embedding approach, inspired by CAM2 \cite{cam2_2021} and DICE \cite{dice2021}, but with a key simplification.

\subsubsection{Causal Network Architecture.} The causal tower is an auxiliary neural network. Unlike typical causal tower implementations that require external popularity signals, impression counts, or exposure logs \cite{cam2_2021}, we made a critical design choice to rely solely on outputs already available in the base architecture. Its input is the candidate embedding concatenated with the transformer encoder's output (a learned profile embedding encoding the user's engagement history).

\textbf{Label construction.} Each training sample receives two binary labels following CAM2 \cite{cam2_2021}: if the candidate is popular and the user engaged with it, it is labeled (pop=1, rel=0); if the candidate is non-popular and the user engaged, it is labeled (pop=0, rel=1); otherwise (pop=0, rel=0). Popularity is defined as the top percentile of positive samples by engagement count, computed over several months of engagement data so that newly trending titles are not penalized for short-term popularity. The assumption is that if a user engaged with a non-popular title, it was likely out of genuine relevance rather than conformity. This framing reduces the debiasing problem to two binary segmentation labels derived from existing engagement-count data.

\textbf{Loss.} The total training loss combines the main task objectives with the auxiliary causal losses:

$$\mathcal{L} = \mathcal{L}_{\text{engagement\_prob}} + \mathcal{L}_{\text{engagement\_length}} + w_{\text{pop}} \cdot \mathcal{L}_{\text{pop}} + w_{\text{rel}} \cdot \mathcal{L}_{\text{rel}}$$

The weighting is asymmetric ($w_{\text{rel}} = 1 - w_{\text{pop}}$), with $w_{\text{pop}}$ set to a small value, making $w_{\text{rel}}$ correspondingly higher. As popular titles are already well-represented in the main task's training signal, this emphasizes the tower to learn representations for relevant long-tail titles. The approach replaces per-sample weight calibration with a single threshold and two scalar loss weights. We note that learning these weights jointly (as in CAM2) did not produce stable results in our setting without external data sources. Fixed weights proved more robust. Additionally, in our experiments, we found consistent popularity debiasing across a range of values, provided $w_{\text{rel}}$ was set sufficiently high (hence $w_{\text{pop}}$ low).

\textbf{Gradient isolation.} Backpropagation from the main task towers (engagement probability and engagement length) is disabled to the causal tower. The causal tower's parameters are updated only by its own auxiliary losses. Without this isolation, the main task's gradients would pull the causal embeddings toward popularity-correlated representations (since popular titles genuinely have higher engagement rates), defeating the purpose of disentanglement.

\textbf{Inference.} At inference time, the causal tower's predicted labels are \textit{not used}. Only its learned embeddings, which capture and isolate the popularity/conformity signal, are passed as additional input features to the main MTL head. This allows the ranking head to be \textit{aware} of popularity effects without being \textit{driven} by them. As an ablation, we zeroed out the causal embeddings at inference, retaining only their training-time gradient contribution to the encoder. Performance only slightly degraded, suggesting the tower also functions as a loss regularizer.

In qualitative evaluation, emphasizing relevance learning had a pronounced effect on profiles with niche or underrepresented viewing histories: rather than defaulting to globally popular titles, the model surfaced content aligned with their specific engagement patterns. As a side effect, popular titles still appeared when genuinely relevant to a profile's history, confirming that the tower does not suppress popularity but rather conditions it on user preference. This validated our design choice to forgo external data sources: the two inputs from the base architecture provided sufficient context to determine whether engagement was preference-driven or conformity-driven. The causal tower improved entity coverage and reduced popularity concentration while maintaining engagement predictions (Section 5).

\section{Experimental Evaluation}

\begin{table}[t]
\centering
\small
\caption{Summary of architectural experiments. All experiments use LLM-derived + ID-based content embeddings, a fixed positive-to-negative ratio, and a short-term sequence length. ``YT'' denotes YouTube-style stacked ReLU heads. Best configuration in bold.}
\label{tab:experiments}
\setlength{\tabcolsep}{3pt}
\begin{tabular}{llccl}
\toprule
Experiment & Attention & Causal & MTL & Result \\
\midrule
\textbf{DRanker (final)} & \textbf{Self-attn} & \textbf{Yes} & \textbf{YT} & \textbf{Best overall} \\
Base (no causal) & Self-attn & No & YT & Lower entity coverage \\
Weighted loss & Self-attn & No & YT & Unstable training loss \\
Causal + weighted & Self-attn & Yes & YT & No gain over causal-only \\
Cross-attention & Cross-attn & Yes & YT & Contextual features dominated \\
PLE & Self-attn & Yes & PLE & Gating saturation \\
DCN & Self-attn & Yes & DCN & Polynomial features hurt \\
Seq length 50 & Self-attn & Yes & YT & Ranker acts as retriever \\
Title projection & Self-attn & No & YT & RPE information lost \\
\bottomrule
\end{tabular}
\end{table}

\subsection{Attention Mechanism} We compared two attention configurations:

\textbf{Self-attention (chosen).} The encoder-only transformer with self-attention over the engagement-history sequence consistently produced the strongest results. Self-attention allows each item in the sequence to attend to all others, learning which historical engagements are most informative for the current ranking context.

\textbf{Cross-attention.} We implemented cross-attention between candidate-item features and profile engagement-history features. Without a self-attention encoder, contextual features dominated the attention weights, leading to degenerate behavior: the model recommended kids' content to all users in mornings regardless of engagement history. With a self-attention encoder, cross-attention between candidate features and each encoded history element \cite{vaswani2017} introduced substantially more parameters without improving performance.

\subsection{MTL Head Architecture}

\textbf{YouTube-style stacked ReLU (chosen).} Simple stacked ReLU layers with hard parameter sharing \cite{youtube2016} performed best. This architecture is well-understood, easy to debug, and produced stable training dynamics. Each task receives its own dedicated ReLU stub branching from the shared backbone, allowing the different output scales of engagement probability and engagement length to be learned independently without disturbing shared-layer representations.

\textbf{Progressive Layered Extraction (PLE) \cite{ple2020}.} PLE is a sophisticated MTL architecture with shared and task-specific expert networks connected via gating. In our experiments, gating networks exhibited polarization \cite{zhao2019_youtube_multitask}: the softmax gating distribution collapsed so that most experts converged to near-zero utilization while a small subset received disproportionate routing. We also observed dying ReLU problems in the expert layers \cite{lu2019dying_relu, shazeer2017moe}, where neurons permanently output zero and stop contributing to learning. Switching to GeLU activations, adding bias terms, and modifying initialization \cite{lu2019dying_relu} only partly mitigated these issues, and the stabilized PLE underperformed the simpler YouTube head given our feature count. PLE's advantage lies in managing tasks that pose the risk of negative transfer; with our two positively correlated tasks (engagement probability and expected engagement length), the additional complexity and continued stabilization effort was not justified for our setting.

\textbf{Deep Cross Networks (DCN) \cite{dcn2017}.} DCN produces polynomial feature interactions, which introduced too many cross-features when combined with contextual features. The result was similar to cross-attention: contextual features overwhelmed individual preferences.

\subsection{Pooling and Sequence Length}

\textbf{Mean pooling vs. projection + concatenation.} We experimented with projecting each item embedding to a smaller latent space and concatenating all items. This performed worse, with losses taking longer to reduce over epochs. Our intuition is that the projected representation produced $n \times d_{\text{proj}}$ features, which may have been too many for the downstream layers to effectively learn from. Additionally, relative-position information encoded in the attention outputs was likely lost in the concatenated projection. Mean pooling is parameter-efficient and maps diverse engagement histories into a common latent space where the downstream MTL head can discriminate effectively.

\textbf{Sequence length.} We tested short, medium, and long-term sequence lengths. At the short-term length, the model behaved as a ranking model, focusing on the user's most recent preferences to determine the order of candidates. At longer sequence lengths, the transformer began acting more like a retriever, prioritizing recall over precision (likely due to mean pooling diluting recency signals across longer sequences). Given our pooling strategy and use case as a precision-focused ranker, the short-term sequence length provided the best trade-off.

\subsection{Causal Tower vs. Weighted Loss} We compared three configurations:

\begin{itemize}
\item \textbf{No debiasing:} Strong relevance and engagement-length metrics, but exhibited high popularity bias (Section 4.3).
\item \textbf{Weighted logistic regression:} Unstable training dynamics due to interaction with stratified sampling; results were uninterpretable for minority strata and showed no debiasing effect on others.
\item \textbf{Causal tower (chosen):} Stable training, improved entity coverage, maintained prediction quality.
\end{itemize}

\begin{table}[t]
\centering
\small
\caption{Offline evaluation --- accuracy vs.\ popularity concentration. AP@10 and Top-50 Overlap (fraction of a user's top-100 recommendations appearing in the platform's daily top-50 titles). Evaluated on 100k daily interactions (US, 10-day holdout). Missing interactions are imputed as not relevant. Higher AP@10 is better; lower Top-50 Overlap indicates less popularity bias. The same variant ranking held for NDCG@k and Recall@k across all cutoffs.}
\label{tab:offline}
\setlength{\tabcolsep}{4pt}
\begin{tabular}{lcc}
\toprule
Variant & AP@10 & Top-50 Overlap \\
\midrule
Self-Attn + No Causal + YT & \textbf{$\sim$0.035} & $\sim$10\% \\
Self-Attn + Causal + YT & $\sim$0.028 & \textbf{$\sim$5\%} \\
Self-Attn + Causal + PLE & $\sim$0.018 & $\sim$1\% \\
Cross-Attn + Causal + YT & $\sim$0.015 & $\sim$4\% \\
\bottomrule
\end{tabular}
\end{table}

\subsection{Variant Selection} Based on offline evaluation of the model configurations described above, we selected four variants for online A/B testing that demonstrated the strongest net performance across varied criteria (ranking accuracy, diversity, and qualitative evaluation). These were variants combining self-attention with the YouTube-style MTL head, with and without the causal tower, using two inference score formulations. Table~\ref{tab:offline} summarizes their offline results.

\subsection{Online A/B Test} We tested four variants in a global A/B test across all supported regions and devices. Table~\ref{tab:online} reports relative lift over control.

\textbf{Result.} The deployed variant, DRanker Tower (combined), achieved a statistically significant +0.48\% lift in our key engagement metric, with control being one of the siloed models ranking the majority of the homepage. At Disney+'s scale, this motivated extending the ranker to additional surfaces powered by more specialized, curated models. A subsequent holdout across several surfaces, with control being models previously powering these surfaces (PLE, SASRec, SRGNN), confirmed the aggregate impact: +1.4\% in our key engagement metric and +0.5\% in daily active users, establishing the model as the unified ranker across the platform. Notably, the causal tower variant achieved the highest online engagement lift despite slightly lower offline AP@10 (Table~\ref{tab:offline}), confirming that offline evaluation undervalues models that deviate from historical patterns. The causal tower variant also achieved the largest reduction in search sessions, indicating users found relevant content directly from homepage rankings.

\begin{table}[t]
\centering
\small
\caption{Online A/B results --- relative lift (\%) over control. \textbf{Bold} = statistically significant ($p < 0.01$). Best per metric highlighted with $\dagger$.}
\label{tab:online}
\setlength{\tabcolsep}{2.5pt}
\begin{tabular}{lcccc}
\toprule
 & Base & Base & Tower & Tower \\
Metric & (prob) & (combined) & (prob) & (combined) \\
\midrule
Engagement/User & \textbf{+0.43} & \textbf{+0.41} & \textbf{+0.35} & \textbf{+0.48}$^\dagger$ \\
Title Diversity & \textbf{+1.02}$^\dagger$ & \textbf{+0.50} & \textbf{+0.84} & \textbf{+0.32} \\
DAU & \textbf{+0.29}$^\dagger$ & +0.13 & \textbf{+0.20} & \textbf{+0.24} \\
\% Sessions w/ Search & $-$35.1 & $-$49.3 & $-$41.8 & \textbf{$-$61.6}$^\dagger$ \\
\bottomrule
\end{tabular}
\vspace{2pt}
\par\noindent{\scriptsize Deployed variant: DRanker Tower (combined).}
\end{table}

\section{Real-Time Serving}

DRanker scores tens of millions of predictions per second in production, processing tens of thousands of requests per second with thousands of candidates each at a median inference latency not more than 50ms, typically much lower. The approximately 7 million parameter model operates within a strict couple-hundred-millisecond end-to-end latency budget for full homepage personalization, allocating a small fraction of the budget to inference while leaving headroom for feature extraction, retrieval, and page composition. This represents a 10x performance improvement at 25\% of the cost compared to CPU-based alternatives, making transformer-based ranking economically viable at scale.

The serving architecture separates concerns between Feature Service (feature extraction and quality validation) and Inference Service (inference). Feature Service extracts features from user requests and normalizes them for model scoring. Inference Service handles inference with TensorRT, achieving sub-25ms latency through several pragmatic optimizations: enforcing uniform input shapes to prevent TensorRT execution plan recompilation, CUDA pinned memory for zero-copy host-GPU transfers, lock-free buffer pools to eliminate allocation overhead, and embedding caching to avoid transmitting expensive vectors over the wire. Fleet isolation ensures experimental models never destabilize production traffic.

Model artifacts are versioned in MLFlow with tag-based selection, tying each model to the code that generated it for reproducibility. Before production deployment, Feature Service validates each candidate model against synthetic personas spanning genre preferences, age groups, and geographic regions using the behavioral testing methodology described in Section 7.4. Models that fail validation are blocked from canary deployment.

\section{Lessons Learned}

\subsection{Invest in the Base, Not the Head} The recommender-systems literature has produced a rich set of complex MTL architectures (PLE, MMoE, DCN, etc.) designed to manage even uncorrelated tasks through sophisticated gating, expert routing, and feature-crossing mechanisms. Our experience suggests that for a ranking model with positively correlated tasks, the quality of the input representation matters more than the complexity of the task head.

The single most impactful change in DRanker's development was moving from shorter training-data windows on prior models that averaged engagement-history embeddings to a transformer encoder over enriched sequences. Combining LLM-derived text embeddings with learned ID-based embeddings allowed the model to function as both a content-based and collaborative filtering system within a single architecture. Not enhancing the MTL head with newer predictions or tasks. Once the base produced rich, contextualized representations, even the simplest MTL head (stacked ReLU layers) performed well. Complex heads like PLE and DCN introduced optimization challenges (gating saturation, dying ReLUs, feature-interaction dominance) that offset their theoretical advantages.

\subsection{Low-Effort Debiasing Beats High-Effort Debiasing} The causal tower was our most impactful ``shortcut.'' Standard approaches to popularity debiasing (IPS weighting, weighted logistic regression, propensity scoring) carry the risk of production drift and require careful calibration that interacts poorly with stratified data.

The causal tower required:

\begin{itemize}
\item Two binary threshold labels derived from existing engagement-count data.
\item A small auxiliary network (<5\% of total parameters).
\item One line of code to stop gradients (gradient isolation).
\item Zero additional data pipelines: the causal tower operates on the candidate embedding and the transformer encoder's output.
\end{itemize}

\subsection{Data Engineering Has Outsized Impact} Three data-engineering decisions had disproportionate impact on ranking quality:

\textbf{First-interaction filtering} reduced the training set by over 60\% while improving ranking metrics.

\textbf{Stratification} was essential for global performance. Without it, the model would overfit to popular titles in high-activity regions. Stratifying across titles, regions, languages, and devices ensured the model saw the full diversity of the catalog. For example, changing playback language would yield relevant updates in recommendations.

\textbf{Random negatives are not weak.} Against our discovery-only positive samples, random catalog negatives provided robust contrastive signal. Dissimilar negatives performed worse (trivially distinguishable), and impression-based negatives were prohibitively expensive to join at scale.

\subsection{Enriched Sequences Require Alignment Verification} Encoding not just \textit{what} the user engaged with but \textit{how} they engaged (device, time, duration) into the sequence made a meaningful difference. The transformer's attention mechanism can learn patterns like ``this user engages with animated content on the TV in the evening and action movies on their phone during commute,'' patterns that are invisible to models operating on content embeddings alone. However, while encoding rich features is straightforward, ensuring that the model meaningfully leverages them is both more difficult and more important. To this end, we performed behavioral alignment checks.

\begin{figure}[t]
\centering
\includegraphics[width=\columnwidth]{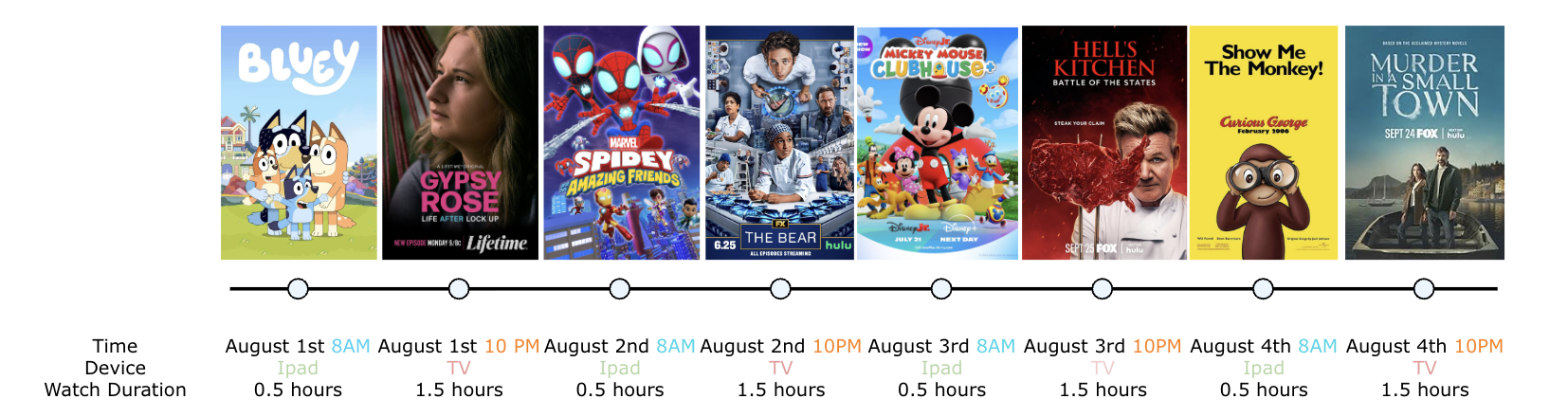}
\vspace{4pt}
\includegraphics[width=\columnwidth]{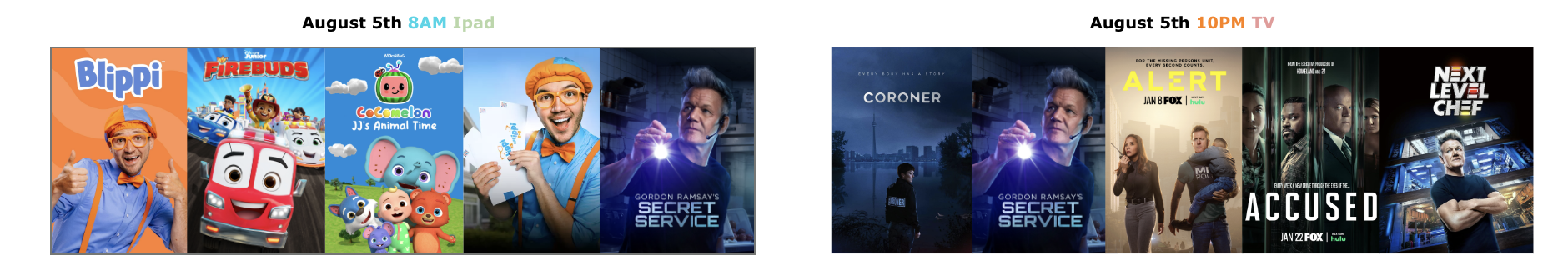}
\caption{Multi-user profile test. Top: synthetic shared engagement history with two distinct patterns (children's content during daytime on mobile, general entertainment during evening on TV). Bottom: the model correctly surfaces children's content for daytime/mobile sessions and general entertainment for evening/TV sessions.}
\Description{Two-panel figure. Top panel: a synthetic shared engagement history with two distinct patterns, children's content engaged during daytime on mobile and general entertainment content engaged during evening on TV. Bottom panel: model recommendations split by session context, surfacing children's content for daytime/mobile sessions and general entertainment for evening/TV sessions.}
\label{fig:multi-user}
\end{figure}

\textbf{Multi-user profiles.} We created synthetic profiles with stereotypical behavioral patterns: a younger user on a mobile device in the morning and an older user on a television in the evening (Figure~\ref{fig:multi-user}). The model correctly differentiated recommendations by session context, surfacing children's content for morning/mobile sessions and adult content for evening/TV sessions, despite both being served from the same engagement history. This confirmed that attention was leveraging device and time-of-day features, not ignoring them.

\begin{figure}[t]
\centering
\includegraphics[width=\columnwidth]{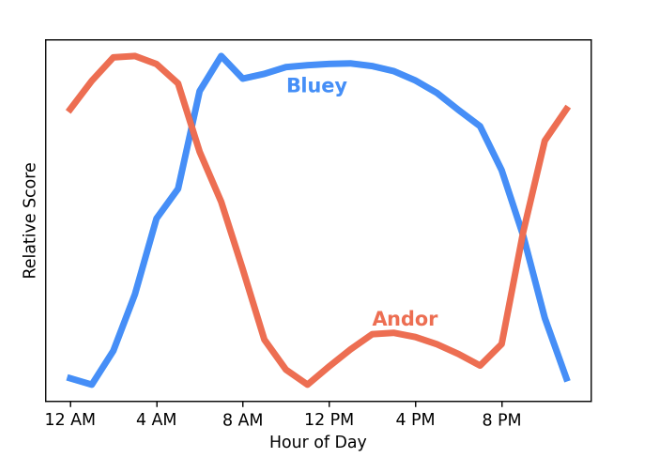}
\caption{Relative model scores by local hour of day. Bluey (children's) peaks during daytime; Andor (mature) peaks in the evening, matching real engagement patterns.}
\Description{Line chart with two curves plotted against local hour of day. Bluey (children's title) scores peak during midday hours. Andor (mature title) scores peak during evening hours.}
\label{fig:title-trends}
\end{figure}

\textbf{Time-of-day scoring.} We examined relative scores produced by the model with respect to local time for key titles. Bluey, a popular children's show, correctly peaked during daytime hours where engagement is highest, while Andor, a Star Wars series targeted towards a more mature audience, showed lower scores during typical working hours (Figure~\ref{fig:title-trends}). This behavior was consistent across a variety of titles on the platform.

This is, in essence, an ML alignment problem: a model can minimize loss without learning the causal structure you intend it to. Qualitative checks like these, which we later systematized into LLM-based evaluation (Section 8), were how we confirmed that enriched features translated into meaningful behavioral differences, not just better loss curves.

\section{Conclusion}

We presented DRanker, the production ranking model for the Disney+ homepage, and the pragmatic engineering process that shaped its design. Our key finding is that simple, well-motivated architectural choices outperformed more complex alternatives in both offline and online evaluation. DRanker is deployed globally, serving hundreds of millions of users and driving a statistically significant lift in our key engagement metric.

\textbf{Future Work.} DRanker represents a starting point. We are actively pursuing several directions:

\begin{itemize}
\item \textbf{Dual evaluation methodology.} Production bias makes it difficult to beat existing systems on offline metrics alone. We are pursuing two complementary directions: (1) \textit{LLM-based evaluation} --- our experience with mock-profile test sets and automated behavioral validation in the model promotion pipeline (Section 6) demonstrated the value of qualitative evaluation for hypothesis verification. We are investing heavily in scaling this approach to systematically set and track ranking objectives. (2) \textit{Counterfactual offline evaluation} --- developing offline policy estimators using doubly-robust methods \cite{dudik2011dr, gilotte2018offline, schnabel2016} to reduce dependence on live A/B tests for model selection and iteration.
\item \textbf{Enhancing data and model foundations.} On the data side, we are incorporating impression-based negatives (shown-but-not-engaged titles) to model implicit negative signals. The session-reactivity that comes with impression signals demands corresponding model capacity and a multi-objective formulation. This motivates investment in the MTL head to balance engagement, diversity, and long-term satisfaction.
\item \textbf{Multi-domain and multi-modal expansion.} Extending the ranker beyond entertainment to other Disney+ content verticals (sports, news), which have fundamentally different engagement patterns and content lifecycles, and incorporating multi-modal inputs (visual, audio) beyond text embeddings.
\end{itemize}

\begin{acks}
We'd like to thank the Disney+ Recommendations team for their support and contributions to the development and productionalization of the DRanker system. In particular we'd like to acknowledge Tom Chee-Duffy, Nicholas Flores, Michael Fullen, Gracey J. Kim, Viral Shah, Annie Cooper, Nate Quinn, Yoav Caspi, Brandon Colton, and Joaquin Delgado.
\end{acks}

\bibliographystyle{ACM-Reference-Format}
\bibliography{references}

\end{document}